\documentclass[runningheads]{llncs}

\usepackage{eccv}

\usepackage{eccvabbrv}

\usepackage{graphicx}
\usepackage{booktabs}
\usepackage{algorithm}
\usepackage{algpseudocode}
\usepackage{soul}
\usepackage{amsmath}

\usepackage[accsupp]{axessibility}  

\usepackage{hyperref}

\usepackage{orcidlink}
\DeclareMathOperator*{\argmin}{arg\,min}
\DeclareMathOperator*{\avg}{avg}

\begin{document}

\title{ECTraj: Enhanced Consistency Training for Multi-Agent Trajectory Prediction} 

\titlerunning{ECTraj}

\author{Alen Mrdovic\inst{1} \and
Qingze (Tony) Liu\inst{1} \and
Danrui Li\inst{1} \and
Mathew Schwartz\inst{1} \and
Kaidong Hu\inst{1} \and
Sejong Yoon\inst{2} \and
Mubbasir Kapadia\inst{1} \and
Vladimir Pavlovic\inst{1}
}

\authorrunning{A.~Mrdovic et al.}

\institute{Rutgers University, New Brunswick \and 
The College of New Jersey
}

\maketitle

\begin{abstract}
   Diffusion models for multi-agent trajectory prediction are limited by iterative denoising, which causes inference latency that hinders their use in time-critical settings like autonomous driving. Fast-sampling variants using DDIM and informed initial noise distribution partially alleviate this issue, but they either fail to achieve true single-step generation or are constrained by the chosen noise distribution. Consistency Models (CMs) offer high-quality one-step generation by mapping noise directly to data, but are difficult to train from scratch. We propose \textbf{\textit{ECTraj}}, an enhanced CM pipeline with improved training and conditional generation for trajectory prediction. Our framework extends the student-teacher consistency training scheme: the student produces standard outputs, while the teacher explicitly fuses its predictions with parts of the ground truth to give stronger supervision. We also exploit CMs’ direct denoising for top-K multi-shot generation during training. Combining conditional generation with this enhanced consistency objective yields faster inference and improved prediction accuracy, establishing competitive new benchmarks on the large-scale Argoverse 2 dataset.\footnote{Code: \hyperlink{https://github.com/am3338/ECTraj}{https://github.com/am3338/ECTraj}}
    
\end{abstract}

\section{Introduction}
Trajectory prediction forecasts the short-term future motion of surrounding traffic participants from their observed history, environmental layout, and social interactions. It is vital for safety-critical applications such as robotic motion planning ~\cite{luo2024physical, samavi2025sicnav,singha2026crowd} and autonomous driving ~\cite{madjid2025trajectory, raina2025trajectory}, where predicted motions are required to be physically plausible, socially acceptable, and reflect multiple potential outcomes of participants' intents.  
Recent work using diffusion models for trajectory prediction ~\cite{jiang2023motiondiffuser,wang2024optimizing, mao2023leapfrog, gu2022stochastic} has shown strong performance in the multi-modal setting and the flexibility to impose various sampling techniques to meet additional output constraints. However, high computational complexity hinders the application of diffusion models, where real-time inference is crucial during deployment in systems such as autonomous vehicles. 

During inference, diffusion models gradually denoise a standard Gaussian sample toward the desired data point by following the dynamics specified by a stochastic differential equation (SDE), a process that typically demands hundreds of optimization steps. More efficient variants, such as DDIM\cite{songdenoising}, speed up sampling and reduce the number of required steps; however, they still need more than 10 denoising iterations.
Beyond accelerated diffusion formulations, previous work \cite{mao2023leapfrog,wang2024optimizing} also tried to replace the standard Gaussian with an informed initial distribution. These methods denoise from a lower-variance noise distribution whose means come from trajectory predictions of pre-trained or standalone modules. This warm start reduces the required diffusion steps and promotes a better exploration of multi-modal futures using additional prior knowledge \cite{li2026agma}.
However, such explicit usage makes the model heavily dependent on the pre-trained predictor, as denoising from an inaccurate prior hinders the model from learning the correct denoising path. 

Consistency Models (CMs)\cite{song2023consistency} enable single-step sampling without relying on prior-informed sampling. CMs are a student-teacher based framework, either distilled from pre-trained diffusion models or trained from scratch, where the first approach achieves strong results in image generation. However, distillation requires robust pre-trained diffusion models, which are found scarce in trajectory prediction domain (shown later in this paper). The latter train-from-scratch method, on the other hand, requires carefully-tailored training guidance to ensure the quality.

We propose \textbf{ECTraj}, a conditional CM-based trajectory prediction framework. ECTraj is conditioned on both historical observations and a pre-trained prior, while performing denoising from a standard Gaussian distribution—thereby maximizing multi-modal exploration without being restricted by an informed initial distribution—and incorporates an enhanced consistency training pipeline. Additionally, the denoising component of ECTraj does not employ consistency distillation~\cite{song2023consistency}, and is instead trained from scratch~\cite{song2023consistency, songimproved}. Our main contributions can be summarized as follows:
\vspace{-\topsep}
\begin{enumerate}
    \item Instead of direct distillation or prior-informed sampling, we guide the training process by first using pretrained model components to encode the input into the latent space and then enhancing the student–teacher training with a ground-truth fusion to the teacher's output.
    \item Tailoring for the multi-modal nature of the trajectory predictions, we exploit one-step denoising to sample multiple modes for student and teacher, compute a best-of-K loss for each network; this optimization is new for CMs and not directly applicable to noise-predicting diffusion models.
    \item ECTraj outperforms the diffusion baseline OptTrajDiff on all standard trajectory benchmark metrics in the large-scale Argoverse 2 dataset, while using only 1/10 of the inference steps.
\end{enumerate}

\section{Related Work}
\subsection{Diffusion Models for Trajectory Prediction}
Diffusion models~\cite{ho2020denoising, song2019generative} have achieved impressive results in image generation. More recently, they have been shown to be a promising framework for trajectory prediction. MID~\cite{gu2022stochastic} captures the uncertainty of agent's future movements by modeling the process as conditional diffusion, where the encoded historical context serves as a condition to generate future trajectories. LED~\cite{mao2023leapfrog} speeds up the sampling process by using DDIM denoising formulation and introducing a more informed prior, which is different from standard Gaussian noise. This in turn reduces the number of sampling steps needed to generate high-quality trajectories. TrajDiffuse \cite{liu2024trajdiffuse} transforms the prediction task into planning-based trajectory inpainting and introduces map-based guidance module to ensure environmental compliance in model output. MotionDiffuser~\cite{jiang2023motiondiffuser} applies diffusion models to the task of multi-agent trajectory prediction for autonomous driving, where they incorporate a permutation-invariant denoiser to enhance multi-agent prediction. OptTrajDiff~\cite{wang2024optimizing} is a multi-agent prediction method that, similarly to LED, forms an informed prior as a via pretrained QCNet~\cite{zhou2023query} and proposed estimated clean manifold guidance to boost the performance in controllable generation. However, all existing approaches still require more than 10 denoising steps, which limits the practicality of deploying the model in time-sensitive applications.
\subsection{Consistency Models}

To alleviate the computational overhead of iterative inference, various fast-sampling diffusion techniques have been developed~\cite{songdenoising, salimansprogressive, karras2022elucidating, song2023consistency}. Prominent among these are Consistency Models (CMs)~\cite{song2023consistency}, which achieve single-step generation from noise via a student-teacher training paradigm. CMs are typically optimized using one of two strategies: Consistency Distillation (CD) or Consistency Training (CT). While CD initializes the model from a pretrained diffusion model, CT trains the model entirely from scratch. Although CD initially outperformed CT by a wide margin, more recent baselines~\cite{songimproved} have systematically refined the design choices for CT, effectively bridging this performance gap. However, due to the lack of standardized pre-trained diffusion baselines for CD training, CMs have yet to be successfully adapted for the trajectory prediction domain. Inspired by recent extensions to the CM framework~\cite{kimconsistency, gengconsistency, fransone, gengmean}, we propose ECTraj, a novel conditional CM-based trajectory prediction pipeline. ECTraj achieves state-of-the-art performance, while requiring only a tenth of the computational time compared to its diffusion counterparts. 

\section{Method}

\begin{figure*}[t]
    \centering
    \includegraphics[width=\linewidth]{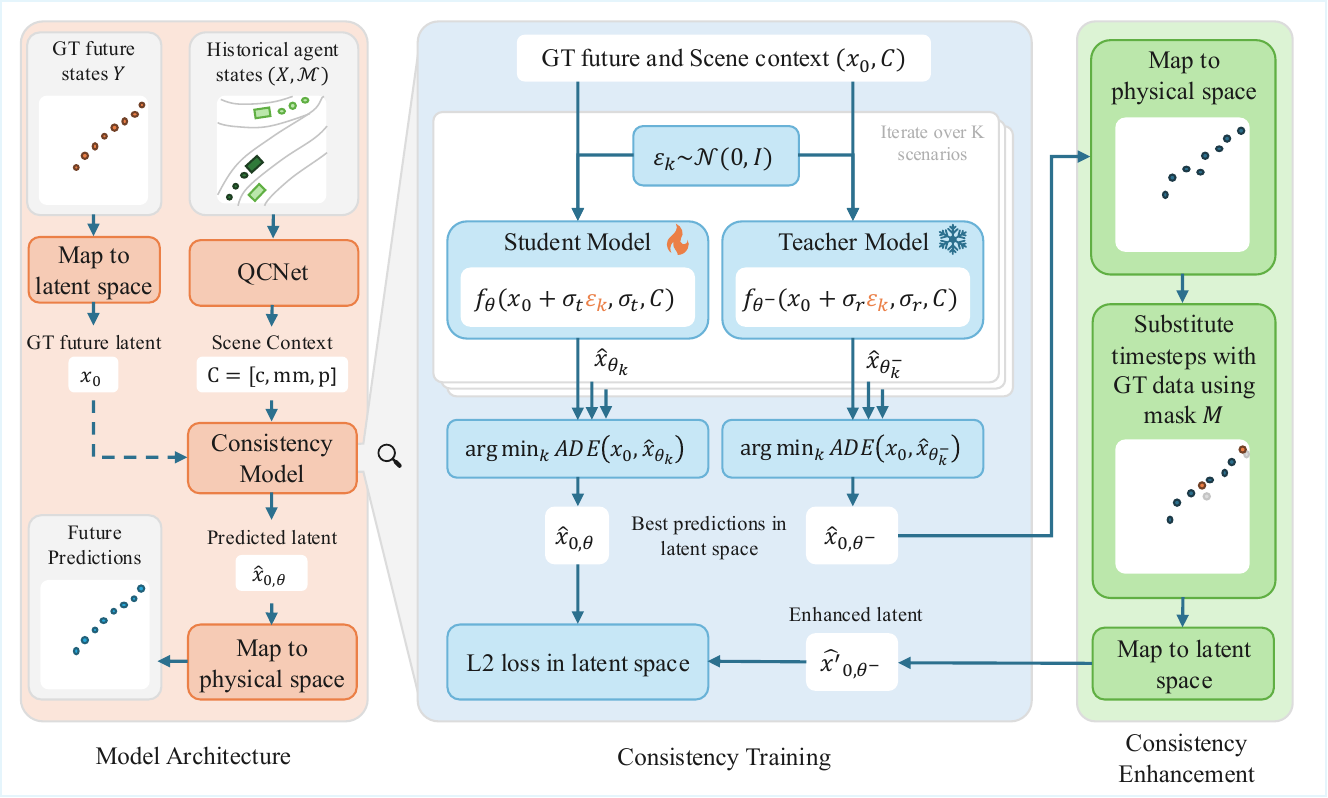}
    \caption{
    \textbf{Model Architecture and training scheme}. (Left) The given historical trajectories and map information are encoded into a context latent vector. Then, the latent is processed by a consistency model and decoded to predicted future trajectories. (Middle) In the teacher-student training scheme of the consistency model, ECTraj samples $K$ different Gaussian noises to produce $K$ different future trajectories. Then the student model is updated using the distance between the best student latent prediction and the best teacher latent prediction. (Right) To enable better teacher guidance, the predicted teacher latent is enhanced by replacing some time steps with GT data in physical space.
    }
    \label{fig:architecture}
\end{figure*}

The complete high-level architecture of our proposed model and its training scheme are depicted in \Cref{fig:architecture}. We describe each component in detail below.

\subsection{Consistency Models Preliminaries}

Consistency models (CMs) are built on the Probability Flow ODE (PF-ODE) in continuous-time diffusion models~\cite{song2021scorebased}, which defines a smooth bijective trajectory that transforms the data distribution \(p_0(x_0)\) into a tractable noise distribution \(p_{\sigma_T}(x_{\sigma_T})\sim \mathcal{N}(0,\epsilon)\). CMs impose a self-consistency constraint so that points \(x_{\sigma_t}\) and \(x_{\sigma_r}\) on the same trajectory map to the same initial data point \(x_0\):
\[
f(x_{\sigma_t},\sigma_t)=f(x_{\sigma_r},\sigma_r)=x_0,
\]
enabling high-quality single-step generation from any intermediate state.\\ \\
For training, the PF-ODE is discretized into time steps \(t_1 < t_2 < \cdots < t_N = T\). The associated variances \(\sigma_1 < \sigma_2 < \cdots < \sigma_N = T\) are obtained as a function of the time step, which is described in more detail in~\Cref{sec:ectraj}. CM then minimizes the distance between the denoised outputs of two adjacent steps:
\[
\arg \min_{\theta} \mathbb{E}\big[w(\sigma_t)\,d\big(f_{\theta}(x_{\sigma_t},\sigma_t), f_{\theta^-}(x_{\sigma_r},\sigma_r)\big)\big],
\]
where \(t_1 \le r < t \le T\). Here \(d(\cdot,\cdot)\) is a distance metric, \(f_{\theta}\) is the neural network that estimates the consistency function, and \(f_{\theta^-}\) is its exponential moving average (EMA).

\subsection{Conditional CM for Multi-modal Trajectory Prediction}

We propose a conditional CM framework for multi-agent trajectory prediction. A scene $(X, \mathcal{M})$ consists of agent states $X \in \mathbb{R}^{N_a \times T_h \times 2}$, where $N_a$ is the number of agents, $T_h$ the number of historical frames, and the last dimension the $(x,y)$ coordinates, and a map $\mathcal{M} \in \mathbb{R}^{N_m \times D_p \times D_m}$, where $N_m$ is the number of polylines, $D_p$ the number of points per polyline, and $D_m$ the number of attributes (e.g., position, orientation, magnitude). Given the encoded scene context $\mathbf{C}$, we learn a conditional consistency function $f_{\theta}(x_{\sigma_t}, \sigma_t, \mathbf{C})$ that directly maps a noisy prior to the future trajectory space to future predictions for $Y \in \mathbb{R}^{N_a \times T_f \times 2}$.
To construct the conditional consistency function $f_{\theta}$, we first encode the historical observations of the surrounding environment and social interactions into a context vector 
\begin{equation}
    c = \Phi(X,\mathcal{M)}.
\end{equation}
To account for the inherently multimodal nature of agent motion, prior work typically introduces auxiliary modeling mechanisms that construct priors—either as deterministic anchor trajectories or as learned prior distributions—to facilitate exploration over multiple plausible futures. 
However, owing to the characteristics of consistency models (CM), which perform denoising directly from Gaussian white noise, it is nontrivial to decode trajectories directly from such priors. Consequently, we instead supply the consistency function with marginal future predictions and their associated scores, obtained from a pre-trained model, as prior anchor trajectory contexts, denoted by $mm$ and $p$ during training time to direct the model with more efficient learning. 
\begin{equation}
    mm,p = \Theta(c)
\end{equation}
After forming the combined context representation $\mathbf{C} = [c, mm, p]$, we apply three layers of composite attention blocks to the intermediate noisy input $x_\sigma$.
Within each layer, we first perform cross-attention between $x_\sigma$ and the historical encodings of the target agent. We then apply cross-attention between $x_\sigma$ and the map encodings (e.g., polygonal representations), followed by cross-attention with the context encodings of the neighboring agents. Subsequently, we conduct self-attention among the future encodings of different agents to capture their joint interactions. Finally, we introduce an additional cross-attention operation between the marginal priors and the noisy input, thereby integrating prior information into the denoising process and obtained the denoising output.
\begin{equation}
    \hat{x}_0=f_{\theta}(x_\sigma,\sigma,\mathbf{C})
\end{equation}
We point out that the consistency function outputs only the (latent) trajectories. The function does not calculate the score for each joint prediction. Since calculating this score can provide useful information about the confidence of each prediction, joint score calculation can be considered in future work.\\ \\
\textbf{Input compression.}
\label{sec:training-details}
In the trajectory space, the flattened input has dimension $N_h \times 2$. Following~\cite{jiang2023motiondiffuser} and~\cite{wang2024optimizing}, we accelerate the decoding by linearly compressing the trajectories into 10-dimensional latent vectors. The trajectory-level input $X$ is mapped to the latent input $x$ as:
\begin{equation}
    x = XU,
\end{equation}
and mapped back by applying:
\begin{equation}
    X = Vx.
\end{equation}
The matrices $U$ and $V$ are learnable; details are given in the Appendix.

\subsection{ECTraj - Enhanced Consistency training}
\textbf{Consistency formulation.}
\label{sec:ectraj}
During training, we first sample the \textit{student} index $t$ from $\{1, \dots, N\}$, where $N$ is the number of discretization steps of the ODE trajectory and determines the noise added to the clean input. Following ~\cite{songimproved}, $N$ is chosen adaptively. In ECTraj, we start with $N = 10$ and, over $E$ training epochs, double the value of $N$ every $\frac{E}{3}$ epochs, so training ends with $N = 40$. The index $t$ is sampled from a discrete log-normal distribution over indices, as in ~\cite{songimproved}; further details are in the Appendix. Given $t$, we then obtain the \textit{teacher} index $r$ as follows:
\begin{equation}
\label{eq:teacher-index}
    r = \lfloor t (1 - \frac{n(t')}{q^{\lceil\frac{e}{d} \rceil}}) \rfloor.
\end{equation}
This parameterization is applied in line with ECT~\cite{gengconsistency}. In~\Cref{eq:teacher-index}, $t' = \frac{t}{N}$, scaling $t$ to $t' \in [0, 1]$, and $n(t') = 1 + \frac{k}{1 + e^{bt'}}$; where $k = 8$ and $b = 1$. Additionally, $e$ denotes the current epoch and $d$ denotes a threshold epoch that is aligned with the discretization curriculum. More precisely, $d = E // 3$ (where $//$ denotes integer division). In cases where $n(t') > q^{\lceil\frac{e}{d}\rceil}$, $r$ is clamped to 0 to avoid negative indexing. The choice of $q$ varies between domains and $q = 4$ was the best choice for our use case. This parameterization leads to a hybrid diffusion-consistency training procedure. In the beginning, the model essentially optimizes for reconstruction, since $r = 0$ in the initial stages of training due to clamping. As training progresses, $r$ approaches $t$, transitioning the training process to standard consistency training~\cite{song2023consistency} as training nears its end. ~\Cref{eq:teacher-index} describes the index retrieval. The noise to be added to the input is a function of the sampled index. For some index $t$, the variance is calculated in line with ~\cite{karras2022elucidating}:
\begin{equation}
    \sigma_t = (\sigma_{min}^{\frac{1}{\rho}} + \frac{t - 1}{N - 1}(\sigma_{max}^{\frac{1}{\rho}} - \sigma_{min}^{\frac{1}{\rho}}))^\rho,
\end{equation}
where $\sigma_{min} = 0.002$, $\sigma_{max} = 1.0$\footnote{This is different from the standard training for image generation where $\sigma_{max} = 80$. This value proved to be too large for our use case.} and $\rho = 7$. For the purposes of discrete log-normal sampling, we also left-pad $\sigma_0 = 0.0$. The explanation for prepending this value is deferred to the Appendix. The student and teacher samples are obtained as:
\begin{gather}
    x_{\sigma_t} = x_0 + \sigma_t\epsilon, \\
    x_{\sigma_r} = x_0 + \sigma_r\epsilon,
\end{gather}
where $\epsilon \sim \mathcal{N}(0, \mathbf{I})$, with $\mathbf{I}$ being the identity matrix. We point out that $x_0$ are compressed trajectory vectors and that optimization is done in the latent space.\\ \\
\textbf{Multi-shot generation at training time.}
Our model produces high-quality trajectories in a single forward pass, which allows us to define a top-$K$ consistency loss. This type of loss is not practical for diffusion-based methods such as OptTrajDiff~\cite{wang2024optimizing}, which operate in the noise space, making it non-trivial to identify the best trajectory. We demonstrate this empirically in~\Cref{sec:ablation-study}. Specifically, we sample $K$ random noise vectors and add them to both the student and the teacher:
\begin{gather}
    x_{\sigma_t, k} = x_0 + \sigma_t\epsilon_k, \\
    x_{\sigma_r, k} = x_0 + \sigma_r\epsilon_k,
\end{gather}
for all $k \in \{1, 2, ..., K\}$. $\epsilon_k \sim \mathcal{N} (0, I)$ represents $K$ different random noise vectors. Next, we separately denoise the trajectories from each group ($K$ student and $K$ teacher modes). After denoising, we choose the best trajectories from both groups based on $ADE$:
\begin{gather}
    k_\theta = \underset{k}\argmin \text{ADE} (f_\theta (x_{\sigma_t, k}, \sigma_t, \mathbf{C}), x_0), \\
    k_{\theta^-} = \underset{k}\argmin \text{ADE} (f_{\theta^-} (x_{\sigma_r, k}, \sigma_r, \mathbf{C}), x_0).
\end{gather}
$\theta^-$ is obtained using the exponential moving average (EMA) of $\theta$. In line with ~\cite{karras2022elucidating}, the denoiser $f_\theta$ is parameterized as follows:
\begin{equation}
    f_\theta(x, \sigma, \mathbf{C}) = c_{skip}(\sigma)x + c_{out}(\sigma)F_\theta(x, \sigma, \mathbf{C}),
\end{equation}
where $F_\theta$ denotes the trainable part of the denoiser (usually a neural network). The teacher parameterization is done similarly. $c_{skip}(\cdot)$ and $c_{out}(\cdot)$ denote differentiable functions for which $c_{skip}(\sigma_{min}) = 1$ and $c_{out}(\sigma_{min}) = 0$. The exact choices of these functions will be provided in the Appendix. For brevity, we will use the following two notations:
\begin{gather}
    \hat{x}_{0, \theta} = f_\theta(x_{\sigma_t,k_\theta}, \sigma_t, \mathbf{C}_t), \\
    \hat{x}_{0, \theta^-} = f_\theta(x_{\sigma_r,k_{\theta^-}}, \sigma_r, \mathbf{C}_r).
\end{gather}
\textbf{Enhanced consistency objective.} Standard CMs directly compare $\hat{x}_{0, \theta}$ and $\hat{x}_{0, \theta^-}$. In ECTraj, we refine the teacher output $\hat{x}_{0, \theta^-}$ by fusing it with parts of the ground truth to provide a stronger supervision for the student. Let $X_f$ be the ground-truth future in trajectory space, and $\hat{X}_{0, \theta^-}$ the predicted trajectory-space output derived from $\hat{x}_{0, \theta^-}$ as described in~\Cref{sec:training-details}. The modified teacher output is then defined as:
\begin{equation}
\label{eq:mixup}
    \hat{X}'_{0, \theta^-} = (1 - M) \odot \hat{X}_{0, \theta^-} + M \odot X_f.
\end{equation}
Here, $M$ denotes a binary mask that determines which parts of the ground truth are replacing the original output, and $\odot$ the Hadamard product. $\hat{X}_{0, \theta^-}$ is then mapped back into the latent space representation $\hat{x}'_{0, \theta^-}$, also as explained in~\Cref{sec:training-details}. We note the similarity between ~\Cref{eq:mixup} and the future mixup in TimeDiff~\cite{shen2023non}, which can also be viewed as a variant of teacher forcing~\cite{williams1989learning}. The key difference lies in masking: TimeDiff uses random masking, while we deterministically replace two waypoints with ground truth—the midpoint and endpoint. Mixing these GT waypoints into the teacher output exposes the student network to higher quality supervision, as this simple modification refines the teacher output and provides the student with several ground truth anchors, thereby aiding learning. The ECTraj objective is:
\begin{equation}
    \mathcal{L}_{ECTraj} = w(\sigma_t) d(\hat{x}_{0,\theta}, \hat{x}'_{0, \theta^-}),
\end{equation}
where $w(\sigma_t) = \frac{1}{\sigma_t - \sigma_r}$ and $d$ is a distance metric (in our case, the $L_2$ norm). \Cref{alg:alg-training} shows the ECTraj training algorithm. The teacher weights are updated as the EMA of the student weights. Since teacher weights are not backpropagated, we apply the stop-gradient operator ($sg$) to them.
\begin{algorithm}[!t]
\small 
\caption{ECTraj Training}
\label{alg:alg-training}
\begin{algorithmic}[1]  
  \State \textbf{Input:} Historical data $(X_h, \mathcal{M})$, ground-truth future data $X_f$, linear mapping $U$ and its inverse $V$, QCNet encoder/decoder $(\Phi, \Theta)$, max epochs $E$, discrete lognormal params $(\mu, \Sigma)$, modes $K$, EMA parameter $\alpha$, binary mask $M$
  \State \textbf{Init:} Randomly initialize $\theta$ and $\theta^-$
  \For{$e = 1,\dots,E$}
    \For{$\big((x_h, m), x_f\big)$ in $\big((X_h, \mathcal{M}), X_f\big)$}
      \State $\mathbf{c} = \Phi(x_h, m)$
      \State $(\mathbf{mm}, \mathbf{p}) = \Theta(\mathbf{c})$
      \State $N = 10 \cdot 2^{\lfloor \tfrac{3e}{E} \rfloor}$
      \State $t \sim \text{DiscreteLogNormal}(\mu, \Sigma, N)$
      \State $r = \lfloor t(1 - \frac{n(t')}{q^{\lceil\frac{e}{d}\rceil}}\rfloor$
      \State $x_0 = x_f U$
      \For{$k = 1,\dots,K$}
        \State $\varepsilon_k \sim \mathcal{N}(0, I)$
        \State $x_{\sigma_t, k} = x_0 + \sigma_t \, \varepsilon_k$
        \State $x_{\sigma_r, k} = x_0 + \sigma_r \, \varepsilon_k$
      \EndFor
      \State $\mathbf{C} = (\mathbf{c}, \mathbf{mm}, \mathbf{p})$
      \State $k_\theta = \underset{k}{\arg\min}\; \text{ADE}\big(f_\theta(x_{\sigma_t, k}, \sigma_t, \mathbf{C}), x_0\big)$
      \State $k_{\theta}^- = \underset{k}{\arg\min}\; \text{ADE}\big(f_{\theta}^-(x_{\sigma_r, k}, \sigma_r, \mathbf{C}), x_0\big)$
      \State $\hat{x}_{0,\theta} = f_\theta(x_{\sigma_t, k_\theta}, \sigma_t, \mathbf{C})$
      \State $\hat{x}_{0,\theta^-} = f_{\theta^-}(x_{\sigma_r, k_{\theta^-}}, \sigma_r, \mathbf{C})$
      \State $\hat{X}_{0,\theta^-} = V x_{0,\theta^-}$
      \State $\hat{X}'_{0, \theta^-} = (1 - M) \odot \hat{X}_{0, \theta^-} + M \odot X_f$
      \State $\hat{x}'_{0, \theta^-} = \hat{X}'_{0, \theta^-} U$
      \State $\mathcal{L}_{ECTraj} = w(\sigma_t)d(\hat{x}_{0,\theta}, \hat{x}'_{0, \theta^-})$
      \State $\theta^- = sg((1 - \alpha)\,\theta + \alpha\,\theta^-)$ \Comment EMA update for teacher; sg = stop-gradient
    \EndFor
  \EndFor
  \State \textbf{return} $\mathcal{L}_{ms}$
\end{algorithmic}
\end{algorithm}

\subsection{Inference}
We design our inference procedure based on the configuration recommended in ~\cite{songimproved}. For multi-step sampling, the intermediate time indices are defined as
\begin{equation}
    \tau_n = \frac{n}{N}\tau_N,
\end{equation}
where $n \in \{1, 2, ..., N\}$. Here, $\tau_N$ is the largest index and corresponds to the maximum variance. ECTraj performs single-step denoising starting from standard Gaussian noise. In contrast to earlier diffusion-based methods~\cite{mao2023leapfrog, wang2024optimizing}, ECTraj does not rely on an informed or specialized initial distribution. Although multi-step inference is feasible, we empirically observed that single-step sampling yields the best performance (multi-step result in supplementary). We hypothesize that this is because multi-step inference excessively corrupts the samples through repeated noise injection, in line with the observations reported in~\cite{song2023consistency}.
\begin{algorithm}
\caption{Inference}
\label{alg:alg-inference}
\begin{algorithmic}
\State \textbf{Input:} Historical data $(X_h, \mathcal{M})$, noisy prior $x_N$, QCNet encoder/decoder $\Phi$/$\Theta$, learned consistency model $\theta$, number of inference steps $N$, intermediate sampling timestep indices $\tau_0 <\tau_1 < \tau_2 < ... < \tau_{N-1}$, such that $\sigma_{\tau_0} = 0$, $\sigma_{\tau_1} = \sigma_{min}$, $\sigma_{\tau_{N-1}} = \sigma_{max}$:
\For{$(x_h, m)$ in $(X_h, M)$}
\State $x_{next} = x_N$
\State $\mathbf{c} = \Phi(x_h, m)$
\State $(\mathbf{mm}, \mathbf{p}) = \Theta(\mathbf{c})$
\State $\epsilon \sim \mathcal{N}(0, I)$
\State $\mathbf{C} = (\mathbf{c}, \mathbf{mm}, \mathbf{p})$
\For{$i$ in N-1..1}
\State $x_{0, \tau_i} = f_\theta(x_{next}, \sigma_{\tau_i}, \mathbf{C})$
\State $x_{next} = x_{0, \tau_i} + \sigma_{\tau_{i-1}}\epsilon$
\EndFor
\EndFor
\State \textbf{return} $x_{next}$
\end{algorithmic}
\end{algorithm}
\Cref{alg:alg-inference} shows the inference algorithm.
\section{Experiments}
\label{sec:experiments}
\textbf{Dataset and metrics. } We conduct our experiments on the Argoverse 2 Motion Forecasting Dataset~\cite{wilson2argoverse}, a widely used large-scale benchmark for trajectory prediction. Each scenario provides 50 past time steps (5 seconds) and 60 future time steps (6 seconds). We adopt the standard metrics from the Argoverse 2 evaluation. Specifically, we incorporate the following accuracy metrics for joint forecasts: $\avg \min ADE_6/FDE_6$ and $\avg \min ADE_1/FDE_6$. In the following text and all tables reporting these metrics, we omit ``$\avg \min$'' to preserve space. $ADE_K$ and $FDE_K$ with $K = 1$ and $K = 6$ to assess the upper bound of the model prediction performance. We further report Brier-$FDE_6$ (abbreviated as $b$-$FDE_6$), which augments the FDE with a confidence penalty term of $(1 - p^2)$, where $p$ is the predicted probability of the best-of-$K$ trajectory. In addition, we measure the miss rate ($MR_K$), defined as the fraction of scenarios in which none of the $K$ predicted trajectories has the $FDE$ less than or equal to $\mathbf{2.0m}$, and the collision rate ($CR_K$), defined as the fraction of scenarios in which at least two agents collide, using a collision distance threshold of $\mathbf{1.0m}$. For $K = 6$, we start with $K$ independently sampled noise vectors. We do not apply any post-processing, as our goal is to evaluate how well ECTraj performs as a standalone consistency-based trajectory prediction model. This is different from OptTrajDiff, which generates $K = 128$ predictions that are used to refine QCNet predictions, using a clustering algorithm to generate candidate joint trajectories. More details are provided in ~\cite{wang2024optimizing}. Another popular multi-agent trajectory prediction dataset is the \textbf{Waymo Open Motion Dataset}~\cite{Ettinger_2021_ICCV}. Results on the Waymo dataset are provided in the Appendix.\\ \\
\textbf{Baselines. } We compared our method with several recent state-of-the-art baselines whose results on Argoverse 2 are publicly available. Specifically, we compare ECTraj to OptTrajDiff~\cite{wang2024optimizing} as its closest counterpart, which is the only joint diffusion-based model directly evaluated on Argoverse 2. We list the benchmark results in~\Cref{tab:marginal-metrics}.\\ \\
\textbf{Training details.} Our model was trained for 60 epochs, using four RTX A6000 GPUs. During training, we used the batch size of 16, and AdamW~\cite{loshchilovdecoupled} optimizer with an initial learning rate of 0.002 and a weight decay of 0.0001. For multi-shot trajectory generation, we generated $K = 6$ future joint trajectories for each scene in the training set, since metric evaluation is commonly done using this number of modes.

\subsection{Quantitative Result: Argoverse 2}

\begin{table}[tb]
\caption{Standard Argoverse 2 metric evaluation}
\label{tab:marginal-metrics}
\centering
\begin{tabular}{@{}llllllllll@{}}
\toprule
Method        & Year & Venue & $ADE_6$ & $FDE_6$ & $ADE_1$ & $FDE_1$ & $b$-$FDE_6$ & $MR_6$ & $CR_6$ \\
\midrule
FJMP~\cite{rowe2023fjmp} & 2023 & CVPR       & 0.81 & 1.89 & 1.52 & 4.00 & 2.59 & 0.19 & \textbf{0.01} \\
GNet~\cite{gao2023dynamic} & 2023 & RA-L       & 0.69 & 1.46 & 1.23 & 3.05 & 2.12 & 0.19 & \textbf{0.01} \\
QCNeXt~\cite{zhou2023qcnext} & 2023 & CVPRW        &    \textbf{0.50}    &   1.02     &   \textbf{0.94}     &   \textbf{2.29}     & \textbf{1.65} & \textbf{0.13}     &  \textbf{0.01}     \\
Forecast-MAE~\cite{cheng2023forecast} & 2023 & ICCV  &    0.69    &   1.55    &    1.30    &   3.33   & 2.24 &   0.19    &  \textbf{0.01}     \\
OptTrajDiff~\cite{wang2024optimizing} & 2024 & ECCV   &    0.60    &   1.31     &   1.08     &  2.71  &  1.95  &  0.17     & \textbf{0.01}      \\
DeMo~\cite{zhang2024decoupling} & 2024 & NeurIPS   &   0.58   &   1.24    &    1.12    &    2.78    & 1.93  & 0.16    &    \textbf{0.01}   \\
RealMotion~\cite{song2024motion} & 2024 & NeurIPS   &  0.62  &   1.32    &   1.14    &     2.87   & 2.01 & 0.18   &   \textbf{0.01}    \\
FutureNet-LOF~\cite{wang2025futurenet} & 2025 & ICRA &    0.58    &   1.25     &   0.96    &\underline{2.34}   &  \underline{1.68} &  0.18     & 0.02      \\
DONUT~\cite{knoche2025donut} & 2025 & ICCV         &    0.55    &   1.13     &   1.07     &  2.62  &  1.79  &  0.15     &  \textbf{0.01}     \\
\textbf{ECTraj (ours)}   & & &  \textbf{0.50 }     &   \textbf{0.96}    &    \textbf{0.94}   &   2.37     & \underline{1.68}  &  \textbf{0.13}    &   \textbf{0.01}   \\
\bottomrule
\end{tabular}
\end{table}

\begin{table}[t]
\caption{Inference speed of ECTraj}
\label{tab:inference-speed}
\centering
\begin{tabular}{lcl}
\toprule
Method   &  NFE & GFLOPs\\
\midrule
OptTrajDiff~\cite{wang2024optimizingA} &  10    & $\sim$ 1311  \\
ECTraj & 1 & $\sim$ 132 \\
\bottomrule
\end{tabular}
\end{table}

\begin{table}[t!]
\caption{Single-scenario inference speed statistics for Argoverse 2 in milliseconds.}
\label{tab:inference-speed-ms}
\centering
\begin{tabular}{@{}lrrr@{}}
\toprule
Method  & Denoising only & QCNet & QCNet + denoising \\
\midrule
QCNet~\cite{zhou2023queryA} & N/A & 24.84 & N/A \\
OptTrajDiff~\cite{wang2024optimizingA} & 29.71 & 24.84 & 54.55\\
ECTraj & 2.97 & 24.84 &27.81\\
\bottomrule
\end{tabular}
\end{table}

Our main experiment consists of comparing ECTraj with several state-of-the-art baselines. The results are shown in~\Cref{tab:marginal-metrics}. ECTraj achieves a state-of-the-art result for $FDE_6$, and is on par with QCNeXt for all metrics. ECTraj achieves better performance than OptTrajDiff across all metrics, showcasing the effectiveness of our CM formulation. Although ECTraj is only the third in terms of $FDE_1$, the runner-up method in terms of this metric (FutureNet-LOF) achieves noticeably weaker accuracy metrics and miss rate for $K = 6$. We also demonstrate additional examples of ECTraj that exhibit better diversity in~\Cref{sec:visuals}.\\ \\ 
\textbf{Inference speed.} One major motivation for replacing the diffusion denoiser with a consistency one lies in its inference speed. We compare our method with OptTrajDiff~\cite{wang2024optimizing} in this aspect and show that our method achieves up to 10 times faster inference compared to OptTrajDiff in~\Cref{tab:inference-speed}. OptTrajDiff uses 10 inference steps, while ECTraj uses only one, which we report in the NFE (\textbf{N}umber of \textbf{F}unction \textbf{E}valuations) column. Since both OptTrajDiff and ECTraj incorporate QCNet into their pipeline, it is useful to evaluate how much computational overhead the denoising component of either method adds. 
In \Cref{tab:inference-speed-ms}, we show the average inference speed per scenario on the Argoverse 2 dataset. We demonstrate that ECTraj adds much less overhead ($\approx 12\%$) compared to OptTrajDiff ($\approx 120\%$). To further demonstrate the ten-fold speedup achieved by ECTraj compared to OptTrajDiff, we also report Giga-floating-point operations (GFLOPs) as a more objective measure compared to raw GPU time. Combined with the superior prediction performance, ECTraj demonstrates the strong potential of using CM as an alternative backbone for the task of trajectory prediction.

\subsection{Qualitative examples}
\label{sec:visuals}
\begin{figure}[t]
    \centering
    \includegraphics[width=1\linewidth]{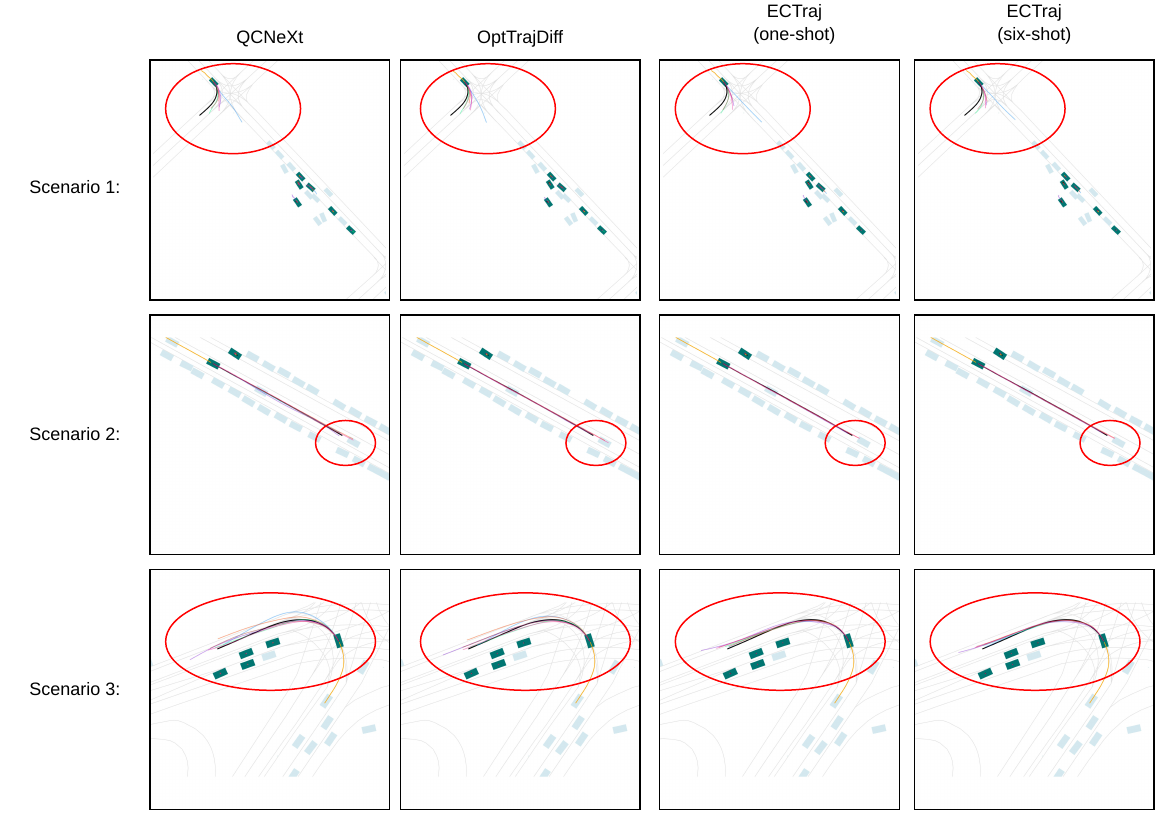}
    \caption{Qualitative examples. Historical trajectories are denoted in \textcolor{orange}{\textbf{orange}}, and ground truth future is denoted in \textbf{black}. Each of the 6 predicted modes is coded in different colors for easier interpretability. Regions of interest in each scenarios are encircled in \textcolor{red}{\textbf{red}}, also for easier interpretability.}
      
    \label{fig:qualitative-examples}
\end{figure}

\textbf{Accuracy, diversity and compliance analysis.}
In~\Cref{fig:qualitative-examples}, we present visual results for the following four methods: \textit{QCNeXt}~\cite{zhou2023qcnext}, \textit{OptTrajDiff}~\cite{wang2024optimizing}, \textit{ECTraj (one-shot loss)} and \textit{ECTraj (six-shot loss)}. In these scenarios, ECTraj excels at various aspects of trajectory prediction:
\begin{itemize}
    \item \textit{\textbf{Scenario 1 - diversity and environmental compliance}}; although all three models are able to model diverse movements, only ECTraj is able to predict an alternate mode that goes straight ahead without violating environmental constraints. Additionally, six-shot ECTraj is able to generate trajectories that better align with the ground truth compared to the one-shot variant.
    \item \textit{\textbf{Scenario 2 - accuracy}}; although all three methods overshoot the ground truth goal point, ECTraj does so to the smallest extent. In this scenario, accuracy-wise, there is no significant difference between one-shot and six-shot ECTraj. This may be due to the simple setting of this scenario, which does not allow for many socially compliant types of intent (i.e. the only feasible intents are going straight ahead or stopping)
    \item \textit{\textbf{Scenario 3 - complex movements}}; although all three methods recognize the intent of a sharp left turn, both QCNeXt and OptTrajDiff predict some modes which pass a non-traversable area, which does not happen in ECTraj. This is even more apparent in the six-shot setting.
\end{itemize}

\begin{figure}[t!]
    \centering
    \includegraphics[width=0.99\linewidth]{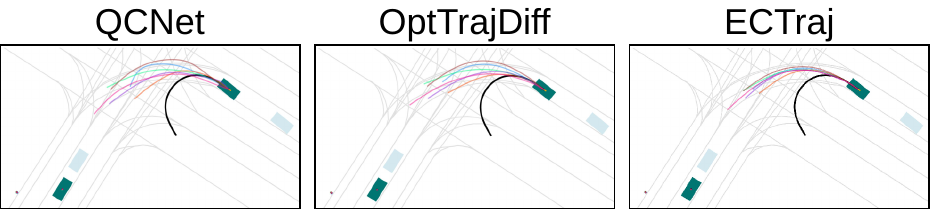}
    \caption{A typical failure mode. Ground truth past shown in \textcolor{orange}{orange}, ground truth future in \textbf{black}, and the six predicted modes in different colors to differentiate them from each other.}
    \label{fig:bad-sample}
\end{figure}

In summary, the qualitative comparisons clearly establish the proposed ECTraj along with the multi-shot loss as the superior approach. Although training with one-shot loss provides a competent baseline for simpler scenarios, it is the six-shot training formulation that unlocks the full robust capability of our method. By improving alignment with ground-truth trajectories and better adherence to environmental constraints during complex maneuvers, ECTraj demonstrate enhanced generative quality with reduced computation latency. 
\\ \\
\textbf{A failure mode: U-turns.}
In addition to scenarios where ECTraj outperforms the baselines such as OptTrajDiff and QCNet, we also demonstrate a typical failure mode of ECTraj, which are U-turns. In \Cref{fig:bad-sample}, we show that this failure mode is not unique to ECTraj and is present in OptTrajDiff and QCNet as well. Additionally, even in this failure scenario, the modes predicted by ECTraj appear to align better with the ground truth mode. We suspect that the main reason for this is the incorporation of the teacher fusion mechanism. This mechanism helps the model determine more easily (although still not entirely successfully) the type of movement it needs to predict, as the agent was stationary prior to making the U-turn, which hinders the inference process.

\subsection{Ablation study}
\label{sec:ablation-study}
In this section, we analyze the impact of enhanced consistency training designs, top-K training in Diffusion formulation, and CM scheduling choices.\\ \\ 
\textbf{Enhanced consistency objective.} 
With the results shown in~\Cref{tab:ablation-study}, all of our training components contribute to the performance of the model. As indicated in row 1 of \Cref{tab:ablation-study}, one-shot training shows reduced performance, which aligns with the observation from qualitative analysis.
We also ablate the teacher output enhancement in two ways. 
\begin{itemize}
    \item \textit{\textbf{Standard CM objective}}: No fusion of ground truth with teacher outputs.
    \item \textit{\textbf{Reconstruction objective}}: We replace every single teacher output point with the ground truth, which reduces the method to the standard direct denoising objective in some variants of diffusion models~\cite{karras2022elucidating}.
\end{itemize}
Both ablations yield suboptimal performance across all metrics. This is intuitive for the following reasons: with the \textit{standard CM objective}, although this objective is the easiest for the student model to learn, it can be unreliable because the teacher model itself may produce inaccurate outputs. With the \textit{reconstruction objective}, the teacher’s output is, in principle, as accurate as possible, but it becomes difficult for the student to learn from—especially when trained with a small number of steps, as in ECTraj. In this case, the student model would require many more iterations to converge. Consequently, the ECTraj training pipeline strikes a careful balance between the quality of supervision and the ease of optimization.
\begin{table}[tb]
\caption{Ablation study on the individual components of ECTraj. The ablated components denote the following: (1) \textit{one-shot training} - replacing six-shot mode generation at training with one-shot, (2) \textit{standard CM objective} - removing fusion of ground truth with teacher, (3) \textit{reconstruction objective} - changing teacher output with ground truth (full ground truth fusion).}
\label{tab:ablation-study}
\centering
\begin{tabular}{@{}llllllll@{}}
\toprule
Ablated component   & $ADE_6$ & $FDE_6$ & $ADE_1$ & $FDE_1$ & $b$-$FDE_6$ & $MR_6$ & $CR_6$\\
\midrule
One-shot training &  0.53      &  1.03      &    0.98     &    2.39         & 1.76 &  0.14   &   0.007    \\
Standard CM objective & 0.52 & 0.99 & 0.96 & 2.38 & 1.70 & 0.14 & 0.007\\
Reconstruction objective & 0.57 & 1.10 & 1.08 & 2.53 & 1.87 & 0.17 & 0.009\\
\textbf{ECTraj} &   \textbf{0.50}     &   \textbf{0.96}     &    \textbf{0.94}     &    \textbf{2.37}    &  \textbf{1.68}   &\textbf{0.13}    &  \textbf{0.006}    \\
\bottomrule
\end{tabular}
\end{table}
\\ \\
\textbf{One-shot vs. multi-shot generation at training time.} As demonstrated in the ablation study above, introducing multiple diverse modes facilitates faster learning, as the model can learn dedicated denoising trajectories for distinct types of motion. In diffusion-based methods such as OptTrajDiff, the best-of-K strategy is not directly applicable: because diffusion models are trained to predict noise, it is not evident how to define or select the “best” noisy mode. Consequently, we investigate two alternative formulations to analyze the impact of multi-shot loss training on the denoising process.
\begin{itemize}
\item \textit{\textbf{Variant 1}}: The best noisy mode is selected as the mode that is the closest to the ground truth \textbf{data}.
\item \textit{\textbf{Variant 2}}: The best noisy mode is selected as the mode that is the closest to the ground truth \textbf{noise}.
\end{itemize}
We studied the performance of these two variants in~\Cref{tab:opttrajdiff-multishot}. Although \textbf{\textit{Variant 1}} does outperform OptTrajDiff in terms of $FDE_6$ and is on par with it in terms of $MR_6$, both variants result in substantially worse metric values compared to ECTraj.\\ \\
\textbf{Consistency scheduling type.} In ECTraj, we use timestep scheduling in line with ECT~\cite{gengconsistency}. In image generation, this method was shown to surpass previous baselines such as \cite{song2023consistency} and \cite{songimproved}. This is also the case in trajectory prediction, as we found that our ECT scheduling achieves a small improvement compared to prior methods (for example, ICT scheduling) \cite{songimproved} in \Cref{tab:ict vs ect}.

\begin{table}[tb]
\caption{Best-of-K approach used in ECTraj tested on OptTrajDiff}
\label{tab:opttrajdiff-multishot}
\centering
\begin{tabular}{@{}llllllll@{}}
\toprule
Variant   & $ADE_6$ & $FDE_6$ & $ADE_1$ & $FDE_1$ & $b$-$FDE_6$ & $MR_6$ & $CR_6$\\
\midrule
Variant 1 &   0.64    &  1.17     &   1.44       &   3.46            & 2.02 &  0.17   &  \textbf{0.01}     \\
Variant 2&  0.99     &  1.42     &    2.38    &   5.10   & 2.30 &  0.21    &   0.03   \\
OptTrajDiff (one-shot) & 0.60 & 1.31 & 1.08 & 2.71 & 1.95 & 0.17 & \textbf{0.01} \\
ECTraj & \textbf{0.50} & \textbf{0.96} & \textbf{0.94} & \textbf{2.37} & \textbf{1.68} & \textbf{0.13} & \textbf{0.01}\\
\bottomrule
\end{tabular}
\end{table}

\begin{table}[tb]
\caption{ICT vs ECT scheduling}
\label{tab:ict vs ect}
\centering
\begin{tabular}{@{}llllllll@{}}
\toprule
Scheduling Type   & $ADE_6$ & $FDE_6$ & $ADE_1$ & $FDE_1$ & $b$-$FDE_6$ & $MR_6$ & $CR_6$\\
\midrule
ICT &   0.52    &   1.02     &   0.95      &   2.38     & 1.77  & 0.15     & \textbf{0.006}     \\
ECT (\textbf{ECTraj}) &   \textbf{0.50}     &   \textbf{0.96}     &    \textbf{0.94}     &    \textbf{2.37}    &  \textbf{1.68}   &\textbf{0.13}    &  \textbf{0.006}    \\
\bottomrule
\end{tabular}
\end{table}

\section{Conclusion and Future Work}
In this paper, we presented \textbf{ECTraj}, a multi-agent trajectory prediction model that utilizes consistency models. We depart from the standard notion of consistency models, and allow fuse parts of the teacher output with the ground truth, where we use the midpoint and the endpoint of the ground truth future to promote accuracy. On top of that, we incorporate a best-of-K training process, which is not feasible in diffusion-based baselines, which optimizes for noise prediction (such as OptTrajDiff). Both of these components help ECTraj achieve state-of-the-art results in almost every Argoverse 2 leaderboard metric, which is particularly noticeable with $FDE_6$. One limitation of ECTraj along with previous diffusion counterparts is its reliance on marginal QCNet priors. We demonstrate the necessity of these priors in the Appendix, showing that the removal of these priors results in worse accuracy metrics on Argoverse 2. That said, in future work, we will consider an end-to-end approach that either does not depend on these priors or uses self-obtained priors.

\bibliographystyle{splncs04}
\bibliography{main}
\end{document}


\title{ECTraj: Enhanced Consistency Training for Multi-Agent Trajectory Prediction - Appendix} 

\titlerunning{ECTraj - Appendix}

\author{Alen Mrdovic\inst{1} \and
Qingze (Tony) Liu\inst{1} \and
Danrui Li\inst{1} \and
Mathew Schwartz\inst{1} \and
Kaidong Hu\inst{1} \and
Sejong Yoon\inst{2} \and
Mubbasir Kapadia\inst{1} \and
Vladimir Pavlovic\inst{1}
}

\authorrunning{A.~Mrdovic et al.}

\institute{Rutgers University, New Brunswick \and 
The College of New Jersey
}

\maketitle
\section{Additional ablation studies}
In this section, we will describe some additional experiments that could not fit in the main submission, which are useful for a more complete assessment of ECTraj.

\subsection{Consistency training hyperparameters}
In our work, we use the time step scheduling in line with~\cite{gengconsistency}. The scheduling itself depends on several hyperparameters. We sample the student index $t$ from the discrete lognormal distribution, which is explained in more detail in~\Cref{sec:lognormal} of this Appendix. Then, we sample the teacher index $r$ as follows:
\begin{equation}
\label{eq:teacher-index}
    r = \lfloor t (1 - \frac{n(t')}{q^{\lceil\frac{e}{d} \rceil}}) \rfloor.
\end{equation}
We define
\[
 n(t') = 1 + \frac{k}{1 + e^{b t'}},\quad t' = \frac{t}{N},\quad t' \in [0,1].
\]
Following \cite{gengconsistency}, we fix \(k = 8\) and \(b = 1\) for all experiments, making \(n(t')\) independent of the dataset.

We set \(d = \lfloor E/3 \rfloor\), where \(E\) is the maximum number of training epochs. This choice is aligned with our discretization curriculum: the number of discretization steps \(N\) doubles every \(\lfloor E/3 \rfloor\) epochs. To keep the schedule consistent, we increase \(d\) in step with \(N\).

The parameter \(q\), in contrast, is dataset-dependent according to \cite{gengconsistency}. Since they evaluate on high-dimensional image data, whereas we work with 10-dimensional latent trajectory vectors, we must tune \(q\) for our setting. We therefore test \(q \in \{2, 4, 6, 8\}\).

\Cref{tab:q-overview} reports the range of the teacher–student index ratio
\[
 \alpha = \frac{r}{t},
\]
for different \(q\) and numbers of discretization steps \(N\) (shown as columns). When \(\alpha < 0\), we clamp it to 0 to avoid negative teacher time-step indices.

\begin{table}[tb]
\caption{Range of teacher-to-student time step ratio $\alpha = \frac{r}{t}$, for each $q$ and each number of discretization steps $N$}
\label{tab:q-overview}
\centering
\setlength{\tabcolsep}{8pt} 
\begin{tabular}{@{}llll@{}}
\toprule
$q$ & $N = 10$ & $N = 20$ & $N = 40$ \\
\midrule
2 &    $\alpha = 0$            &   $\alpha \in [0, 0.21]$              &    $\alpha \in [0.38, 0.60]$             \\
4 &    $\alpha \in [0, 0.21]$             &     $\alpha \in [0.69, 0.80]$             &     $\alpha \in [0.92, 0.96]$             \\
6 &    $\alpha \in [0.17, 0.48]$             &   $\alpha \in [0.86, 0.91]$               &   $\alpha \in [0.98, 0.99]$               \\
8 &    $\alpha \in [0.38, 0.60]$             &    $\alpha \in [0.92, 0.96]$              &      $\alpha \in [0.98, 0.99]$            \\
\bottomrule
\end{tabular}
\end{table}

\Cref{tab:ablation-q} presents an ablation study over the four \(q\) values. While \(q = 2\) performs reasonably well (especially on \(\mathrm{FDE}_6\) and \(b\)-\(\mathrm{FDE}_6\)), \(q = 4\) consistently yields the best overall results. For larger \(q\), performance degrades across metrics. Consequently, we adopt \(q = 4\) in all main experiments.
\begin{table}[tb]
\caption{Ablation on the q time step scheduling hyperparameter}
\label{tab:ablation-q}
\centering
\setlength{\tabcolsep}{8pt} 
\begin{tabular}{@{}llllllll@{}}
\toprule
$q$        & $ADE_6$ & $FDE_6$ & $ADE_1$ & $FDE_1$ & $b$-$FDE_6$ & $MR_6$ & $CR_6$ \\
\midrule
2                   & 0.51 & 0.96 & 1.01 & 2.44 & 1.68 & 0.14 & 0.007\\
4                & \textbf{0.50}  & \textbf{0.96} & \textbf{0.94}  & \textbf{2.37} & \textbf{1.68} & \textbf{0.13} & \textbf{0.006}\\
6                   & 0.52 & 1.04 & 0.99 & 2.50 & 1.77 & 0.14 & 0.007\\
8                & 0.54 & 1.09 & 1.01 & 2.47 & 1.82 & 0.15 & 0.007\\
\bottomrule
\end{tabular}
\end{table}
\subsection{Ground truth fusion}
In the main paper, we tested our final choice of ground truth fusion (midpoint + endpoint) against two other methods. Here, we will provide a more detailed overview of two more variants of fusion we tested our model with. We will provide an overview of each type of fusion here, followed by a corresponding abbreviation used in~\Cref{tab:ablation-gt-fusion} for readability:
\begin{enumerate}
    \item \textit{\textbf{No fusion (NF): }} Fusion is not applied to the teacher output.
    \item \textit{\textbf{Full fusion (FF): }} The teacher output is fully replaced with ground truth - effectively yielding a direct denoising reconstruction objective similar to~\cite{karras2022elucidating}.
    \item \textit{\textbf{Random 2 (R2): }} Random two ground truth points are selected to be fused with the teacher output. This is done in the spirit of TimeDiff~\cite{shen2023non}, where a random binary mask is also used.
    \item \textit{\textbf{Progressive (Prog): }} We start by fusing more evenly spaced points at the beginning of training, after which we gradually decrease the number of points (which are still evenly spaced) to fuse as the training progresses. In our case, we start by replacing \textbf{10} points, and decrease the number of points by \textbf{2} for every 10 epochs.
    \item \textit{\textbf{Midpoint + Endpoint (M+E) (ECTraj): }} We only fuse the midpoint and the endpoint over the entire duration of training.
\end{enumerate}
~\Cref{tab:ablation-gt-fusion} gives an overview of each of these methods. Although some methods are very close to our final version of ECTraj, we see that the \textit{\textbf{M+E}} fusion still gives the best results across all metrics. This may be due to this method's best interpolation between the two extremes:
\begin{enumerate}
    \item \textbf{\textit{Full fusion}}: Although in this case the ``output'' the loss is measured against is the ground truth, which is the most accurate output, learning this objective can be hard, since it requires huge traversals through the PF ODE~\cite{song2023consistency}, which can be difficult at the initial stages of training.
    \item \textbf{\textit{No fusion}}: Although this objective is the easiest to learn due to the proximity of the outputs, it may be inaccurate, since it depends on the teacher model which in itself may be inaccurate, especially at the early stages of training~\cite{song2023consistency, songimproved}.
\end{enumerate}

\begin{table}[tb]
\caption{Ablation on different types of ground truth fusion}
\label{tab:ablation-gt-fusion}
\centering
\setlength{\tabcolsep}{4pt} 
\begin{tabular}{@{}llllllll@{}}
\toprule
Method        & $ADE_6$ & $FDE_6$ & $ADE_1$ & $FDE_1$ & $b$-$FDE_6$ & $MR_6$ & $CR_6$ \\
\midrule
NF                   & 0.52 & 0.99 & 0.96 & 2.38 & 1.70 & 0.14 & 0.007\\
FF                & 0.57 & 1.10 & 1.08 & 2.53 & 1.87 & 0.17 & 0.007\\
R2                    & \textbf{0.50} & 1.00 & 0.98 & 2.47 & 1.70 & 0.14 & 0.007\\
Prog                 & 0.51 & 1.05 & 0.96 & 2.39 & 1.73 & 0.15 & \textbf{0.006}\\
M+E (ECTraj)             & \textbf{0.50}  & \textbf{0.96} & \textbf{0.94}  & \textbf{2.37} & \textbf{1.68} & \textbf{0.13} & \textbf{0.006}\\
\bottomrule
\end{tabular}
\end{table}
\subsection{Multi-step sampling}
In the main paper, we use single-step sampling for all reported ECTraj results. Here, we tested ECTraj with multiple sampling steps, which can sometimes improve metrics in the image domain~\cite{song2023consistency, songimproved}. However, in our case, the best results are achieved with a single sampling step. In our lower dimensional setting, compared to the image domain, successive noise addition may overly corrupt the samples and in turn compromise their quality, which may be even more evident in low-dimensional data. We report this experiment in~\Cref{tab:multistep-sampling}.
\begin{table}[tb]
\caption{Ablation on the number of sampling steps. \textit{NFE} denotes the number of sampling steps}
\label{tab:multistep-sampling}
\centering
\begin{tabular}{@{}llllllll@{}}
\toprule
NFE        & $ADE_6$ & $FDE_6$ & $ADE_1$ & $FDE_1$ & $b$-$FDE_6$ & $MR_6$ & $CR_6$ \\
\midrule
1                   & \textbf{0.50}  & \textbf{0.96} & \textbf{0.94}  & \textbf{2.37} & \textbf{1.68} & \textbf{0.13} & \textbf{0.006} \\
2                 & 0.52 & 1.01 & 0.96 & 2.41 & 1.74 & \textbf{0.13} & 0.007\\
4                   & 0.54  & 1.03 & 0.96 & 2.42 & 1.77 & 0.14 & 0.007\\
\bottomrule
\end{tabular}
\end{table}

\subsection{QCNet marginals as priors}
\begin{figure}[htbp]
\centering
\begin{subfigure}{0.8\textwidth}
  \centering
  \includegraphics[width=.9\linewidth]{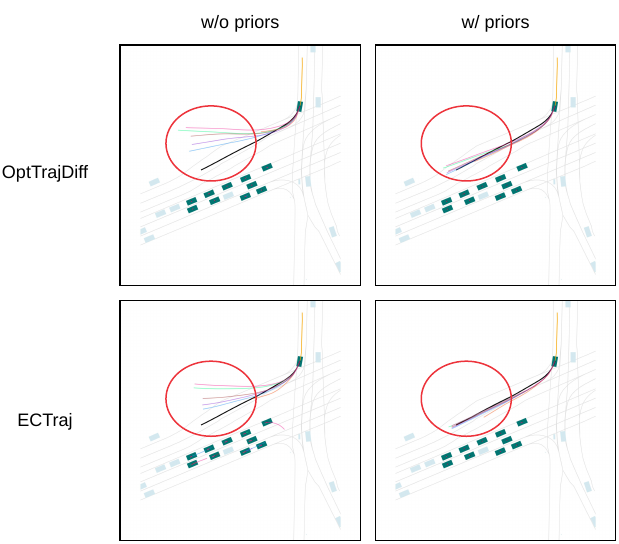}
  \caption{Scenario 1}
  \label{fig:scenario-1}
\end{subfigure}

\begin{subfigure}{0.8\textwidth}
  \centering
  \includegraphics[width=.9\linewidth]{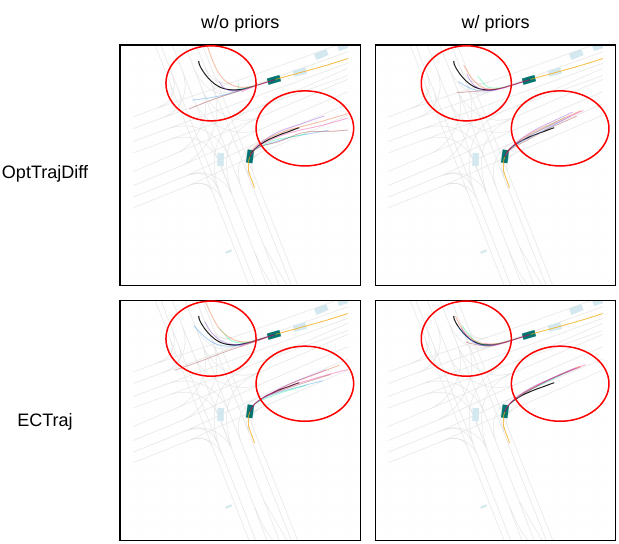}
  \caption{Scenario 2}
  \label{fig:scenario-2}
\end{subfigure}
\caption{Scenarios which benefit from incorporating the QCNet marginals prior. Regions of interest are marked in \textcolor{red}{red}. In both scenarios, not incorporating the prior leads to some modes violating environmental constraints (i.e., driving in the non-driveable area) and, also, decreased accuracy. With the prior introduced, ECTraj gives a more accurate prediction compared to OptTrajDiff, particularly in Scenario 1 (a).}
\label{fig:prior-visuals}
\end{figure}

ECTraj, similarly to OptTrajDiff~\cite{wang2024optimizing}, uses QCNet marginals as priors to facilitate the conditioning process. Here, we ablate the QCNet marginals to observe how well the model performs without them. The QCNet encoder is still used to retrieve the scene encoding.~\Cref{tab:marginal-ablation} shows the ablation study on using marginal QCNet predictions as priors. For both OptTrajDiff and ECTraj, marginals are useful to refine predictions. This is especially noticeable with ECTraj, where the improvement in metrics is relatively higher when marginals are applied. One possible reason for this is the single-step sampling nature of ECTraj - more precisely, priors are more informative when we sample for fewer steps (such as one) compared to more steps (such as 10, like in OptTrajDiff).~\Cref{fig:prior-visuals} shows some scenarios where introducing the QCNet marginals into conditioning significantly improves predictions.
\begin{table}[tb]
\caption{Ablation on QCNet marginals as priors}
\label{tab:marginal-ablation}
\centering
\begin{tabular}{@{}llllllll@{}}
\toprule
Method       & $ADE_6$ & $FDE_6$ & $ADE_1$ & $FDE_1$ & $b$-$FDE_6$ & $MR_6$ & $CR_6$ \\
\midrule
OptTrajDiff                   & 0.68 & 1.36 & 1.37 & 3.60 & 1.99 & 0.20 & 0.02 \\
OptTrajDiff + priors                 & 0.60  & 1.31 & 1.08  & 2.71 & 1.95 & 0.17 & \textbf{0.01}\\
\midrule
ECTraj &  0.64 & 1.34 & 1.52 & 4.14 & 2.03 & 0.19 & 0.02 \\
ECTraj + priors & \textbf{0.50}  & \textbf{0.96} & \textbf{0.94}  & \textbf{2.37} & \textbf{1.68} & \textbf{0.13} & \textbf{0.01}\\
\bottomrule
\end{tabular}
\end{table}

\section{Consistency training details}
In this section, we will discuss some technical details of consistency training.

\subsection{Linear mapping}
For faster training convergence, $120$-dimensional (flattened) trajectories are compressed into $10$-dimensional latent vectors in line with ~\cite{jiang2023motiondiffuser} and ~\cite{wang2024optimizing}. This is done using linear mapping, where the trajectory vectors $\mathbf{x}$ are mapped into latents $\mathbf{z}$ as:
\begin{equation}
    \mathbf{z} = \mathcal{G}(x) = \mathbf{x}U.
\end{equation}
The latents are mapped back to the trajectory space as:
\begin{equation}
    \mathbf{x} = \mathcal{F}(z) = \mathbf{z}V.
\end{equation}
Both $U$ and $V$ are learnable matrices, whose values are obtained by optimizing a composite loss function:
\begin{enumerate}
    \item \textit{\textbf{Reconstruction loss}} - which minimizes the distance between the ground truth trajectory and its reconstruction:
    \begin{equation}
        \mathcal{L}_{rec} = ||\mathcal{F}(\mathcal{G}(X)) - x||^2_2
    \end{equation}
    \item \textit{\textbf{Regularization}} - which aims to make the mappings distance-preserving; in other words, smaller latent-space distances should translate to smaller trajectory-space distances and vice versa. This is done as follows:
    \begin{equation}
        \mathcal{L}_{reg} = (\mathbf{x}^T\mathbf{x} - \mathcal{G}(\mathbf{x})^T\mathcal{G}(\mathbf{x}))^2
    \end{equation}
    \item \textit{\textbf{Variance minimization}} - which aims to reduce training instability due to extreme variances in some dimensions:
    \begin{equation}
        \mathcal{L}_{var} = ||std(\mathcal{G}(\mathbf{x})) - \mathbf{v}||^2_2,
    \end{equation}
    where $\mathbf{v}$ is a vector whose elements are all $\eta$, which is a learnable parameter.
\end{enumerate}

\subsection{Lognormal time step distribution}
During training, the student index $t$ is sampled from the discrete lognormal time step distribution in the main paper. This distribution is formed in line with~\cite{songimproved}, and we will provide some additional details here.\\ \\
\begin{figure*}[t]
    \centering
    \includegraphics[width=0.7\linewidth]{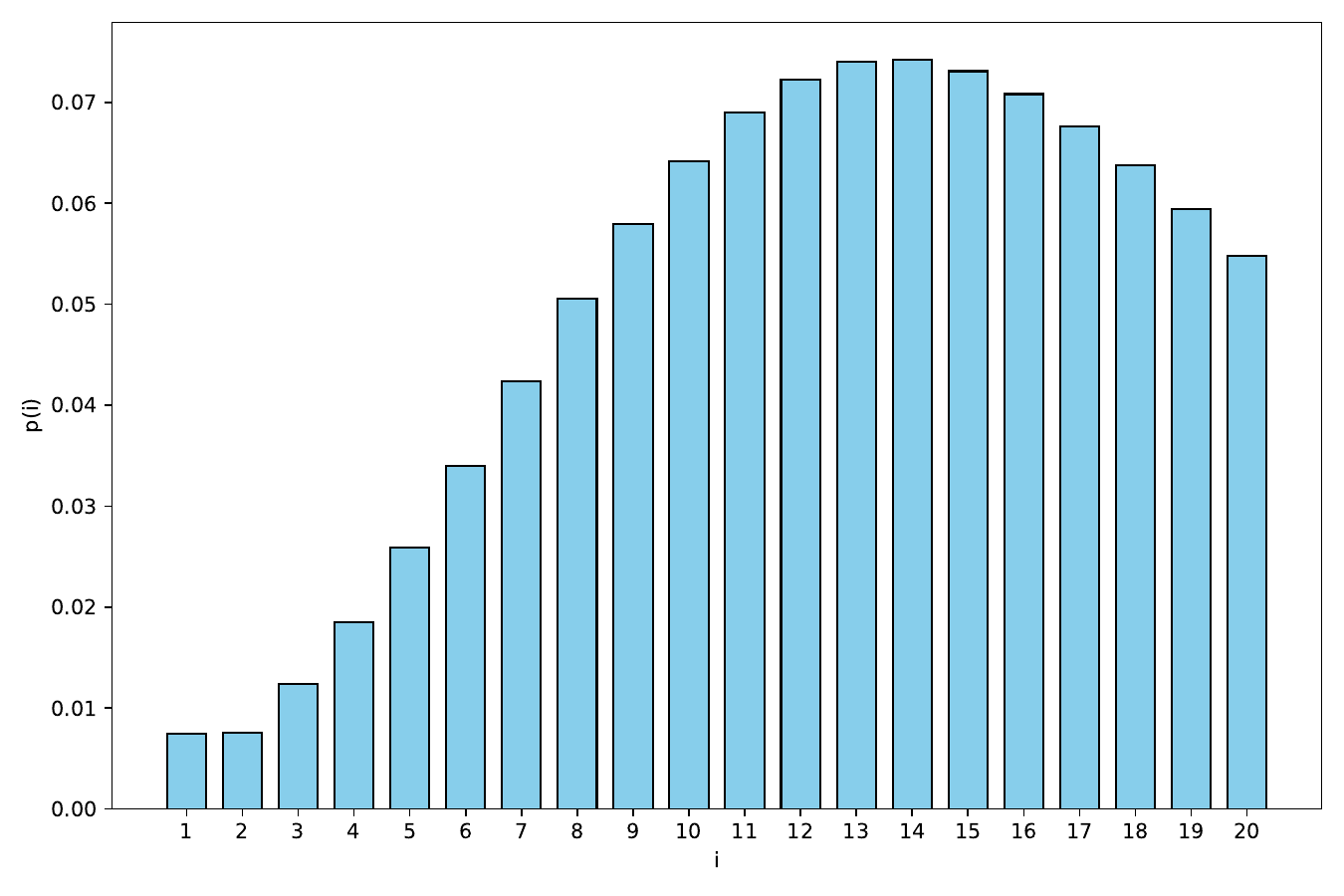}
    \caption{Discrete lognormal time step distribution for $N = 20$}
      
    \label{fig:lognormal-20}
\end{figure*}
\label{sec:lognormal}
First, we recall that each time step index $i$ is associated with a noise variance $\sigma_i$. Taking this into consideration, the discrete lognormal distribution the variance is sampled from can be formulated as:
\begin{equation}
\label{eq:lognormal-pmf}
    ln(\sigma_i) \sim \mathcal{N}(\mu, \Sigma, N),
\end{equation}
where $\mu$ and $\Sigma$ are the mean and variance of the distribution. In our case $\mu = -1.1$ and $\Sigma = 2.0$. $N$ denotes the number of PF ODE discretization steps, which is described in the main paper. Based on~\Cref{eq:lognormal-pmf}, the distribution of variances $\sigma_i$ (and with that, time steps $i$) is defined as:
\begin{equation}
\label{eq:lognormal-sigma}
    p(i) = p(\sigma_i) = \erf \frac{ln(\sigma_{i}) - \mu}{\sqrt{2}\Sigma} - \erf \frac{ln(\sigma_{i-1}) - \mu}{\sqrt{2}\Sigma},
\end{equation}
where $\erf$ denotes the Gauss error function and $i \in {0, 1, ..., N}$. We note that obtaining $p(\sigma_1)$ would require us to additionally define $\sigma_0$, which is simply assigned the value 0 (that is, $\sigma_0 = 0$). $\sigma_1 = \sigma_{min} = 0.002$ and $\sigma_N = \sigma_{max} = 1.0$. All the other values $t \in [2, N-1]$ are obtained in line with~\cite{karras2022elucidating}:
\begin{equation}
    \sigma_t = (\sigma_{min}^{\frac{1}{\rho}} + \frac{t - 1}{N - 1}(\sigma_{max}^{\frac{1}{\rho}} - \sigma_{min}^{\frac{1}{\rho}}))^\rho,
\end{equation}

Since we discretize the PF ODE trajectory during training, we need to sample time steps as integers. We visualize the discrete lognormal distribution for $N = 20$ (second stage of training: epochs 21-40) in~\Cref{fig:lognormal-20}.

\subsection{Consistency parameterization}
As is mentioned in the main paper, the denoiser $f_\theta$ is parameterized as follows:
\begin{equation}
    f_\theta(\mathbf{x}, \sigma, \mathbf{C}) = c_{skip}(\sigma)\mathbf{x} + c_{out}(\sigma)F_{\theta}(\mathbf{x}, \sigma, \mathbf{C}),
\end{equation}
where $\mathbf{x}$ denotes the input, $\sigma$ denotes the noise level (variance), and $\mathbf{C}$ is the condition which consists of the scene context $c$, marginal QCNet predictions $mm$, and their scores $p$. $F_\theta$ denotes the learnable part of our denoiser. This parameterization needs to satisfy the boundary condition:
\begin{equation}
    f_\theta(\mathbf{x}, \sigma_{min}, \mathbf{C}) = \mathbf{x}.
\end{equation}
To accomplish this, in line with~\cite{song2023consistency}, $c_{skip}(\cdot)$ and $c_{out}(\cdot)$ are designed as differentiable functions for which $c_{skip}(\sigma_{min}) = 1$ and $c_{out}(\sigma_{min}) = 0$. We opted for the following choices, similarly to ~\cite{song2023consistency}:
\begin{equation}
    c_{skip} = \frac{0.25}{(\sigma - \sigma_{min})^2 + 0.25},
\end{equation}
\begin{equation}
    c_{out} = \frac{0.5(\sigma - \sigma_{min})}{\sqrt{0.25 + \sigma^2}},
\end{equation}
where $\sigma_{min} = 0.002$.

\section{Results on the Waymo Open Motion Dataset}
Another popular autonomous driving dataset is the Waymo Open Motion Dataset~\cite{Ettinger_2021_ICCV} (which will be referred to as "Waymo" in the following text for brevity). The structure of this dataset is similar to Argoverse 2 Motion Forecasting dataset. The historical time window of Waymo is \textbf{11} frames, whereas the future time window is \textbf{80} frames (cf. Argoverse 2 where the historical time window is \textbf{50} frames and the future time window is \textbf{60} frames).
\subsection{Accuracy metrics}
Since QCNet~\cite{zhou2023query} reports the results on the Waymo dataset, we decided to test ECTraj on that dataset as well. However, one issue we encountered was the lack of the public QCNet checkpoint on the Waymo dataset. For this reason, we had to train QCNet on Waymo from scratch. We have not been able to reproduce the Waymo results from the official report~\cite{zhou2023query}, but we still report the results obtained on Waymo in~\Cref{tab:waymo-results}. The results presented in the table emphasize the key limitation of our method, i.e., the reliance on QCNet priors. This motivates us to explore approaches which do not rely on pretraining in the future.

\begin{table}[tb]
\caption{Results on the Waymo dataset}
\label{tab:waymo-results}
\centering
\begin{tabular}{@{}lll@{}}
\toprule
Method       & $ADE_6$ & $FDE_6$ \\
\midrule
QCNet (official)~\cite{zhou2023query}                   & 0.54 & 1.07   \\
QCNet (our run)                 & 0.77  & 1.62 \\
OptTrajDiff & 0.95 & 2.02 \\
ECTraj  & 0.83 & 1.75 \\
\bottomrule
\end{tabular}
\end{table}

\subsection{Inference speed}
One major motivation for introducing ECTraj in the first place was its improved inference speed. So, it is also important to test this aspect of ECTraj on the Waymo dataset (although it is expected to be dataset-independent). In~\Cref{tab:waymo-inference-speed} we show the average inference speed per scenario, in milliseconds (ms). We see that the computational overhead added by ECTraj is approximately only $8.9\%$, whereas for OptTrajDiff it is approximately $89\%$. This provides additional support for the following two important inference speed claims:
\begin{enumerate}
    \item The computational overhead added by ECTraj is small compared to the compute it takes for QCNet to generate marginal modes, regardless of the dataset
    \item The denoising speed gain of ECTraj is approximately ten-fold compared to OptTrajDiff
\end{enumerate}
Another point that needs to be addressed is the computational overhead for both ECTraj and OptTrajDiff being significantly (around $30\%$) less on Waymo than it is on Argoverse 2, as is reported in the main paper. We believe that one major factor is that QCNet operates in the trajectory space, whereas both ECTraj and OptTrajDiff operate in the latent space. The trajectory space dimensionality varies across datasets ($60 \times 2 = 120$ for Argoverse2, and $80 \times 2 = 160$ for Waymo). However, in both OptTrajDiff and ECTraj, we decided to keep the latent dimension as 10 in both Argoverse 2 and Waymo. This leads to the compression factor of the linear mapping being greater for Waymo, which can in turn lead to more efficient computation.

\begin{table}[tb]
\caption{Single-scenario inference speed statistics for Waymo (ms).}
\label{tab:waymo-inference-speed}
\centering
\begin{tabular}{@{}llll@{}}
\toprule
Method       & Denoising only & QCNet & QCNet + denoising \\
\midrule
QCNet (our run)   &     N/A         & 40.82  & N/A \\
OptTrajDiff & 36.68 & 40.82   & 77.50\\
ECTraj  & 3.67 & 40.82   & 44.49\\
\bottomrule
\end{tabular}
\end{table}

\newpage
\bibliographystyle{splncs04}
\bibliography{main}